# SKETCH: Structured Knowledge Enhanced Text Comprehension for Holistic Retrieval

**Aakash Mahalingam[1], Vinesh Kumar Gande[1], Aman Chadha[2,3,†], Vinija Jain[2,4, ‡], Divya Chaudhary[1]**

[1]Northeastern University, [2]Stanford University, [3]Amazon AI, [4]Meta

mahalingam.aa@northeastern.edu, gande.vi@northeastern.edu, hi@aman.ai, hi@vinija.ai, d.chaudhary@northeastern.edu

## Abstract

Retrieval-Augmented Generation (RAG) systems have become pivotal in leveraging vast corpora to generate informed and contextually relevant responses, notably reducing hallucinations in Large Language Models. Despite significant advancements, these systems struggle to efficiently process and retrieve information from large datasets while maintaining a comprehensive understanding of the context. This paper introduces SKETCH, a novel methodology that enhances the RAG retrieval process by integrating semantic text retrieval with knowledge graphs, thereby merging structured and unstructured data for a more holistic comprehension. SKETCH, demonstrates substantial improvements in retrieval performance and maintains superior context integrity compared to traditional methods. Evaluated across four diverse datasets: QuALITY, QASPER, NarrativeQA, and Italian Cuisine—SKETCH consistently outperforms baseline approaches on key RAGAS metrics such as answer_relevancy, faithfulness, context_precision and context_recall. Notably, on the Italian Cuisine dataset, SKETCH achieved an answer relevancy of 0.94 and a context precision of 0.99, representing the highest performance across all evaluated metrics. These results highlight SKETCH's capability in delivering more accurate and contextually relevant responses, setting new benchmarks for future retrieval systems.

## 1 Introduction

Large Language Models (LLMs), despite their growing size and capabilities, often lack sufficient domain-specific knowledge for certain tasks [27], and their encoded facts can quickly become outdated due to the dynamic nature of information. Updating the knowledge within LLMs through fine-tuning or editing is a complex and resource-intensive process, especially when dealing with extensive text corpora [32]. An alternative approach is Retrieval Augmented Generation (RAG) [45], which involves indexing large volumes of text, divided into smaller segments, in a separate information retrieval system [12]. Retrieved information is then provided to the LLM along with the query as context, enabling more factually updated answers [6]. This method offers benefits such as access to current, domain-specific knowledge, improved interpretability, and provenance tracking, which are often lacking in the opaque parametric knowledge of LLMs [47].

However, existing RAG methods have notable limitations [17]. They typically retrieve only a few short, contiguous text segments, limiting their ability to represent and leverage large-scale discourse structures. This limitation is particularly problematic for complex queries that require integrating knowledge from multiple parts of a text, such as synthesizing information across chapters or drawing conclusions from dispersed data in scientific literature. Moreover, these methods often struggle with multi-hop reasoning, where answering a query necessitates combining information from multiple, non-adjacent text segments, leading to incomplete or inaccurate answers and reducing their usefulness in applications that demand comprehensive understanding and synthesis.

To address these challenges, we introduce SKETCH, a novel methodology that enhances the retrieval process in RAG systems by integrating semantic text retrieval with knowledge graphs [16; 30]. This integration merges structured data (knowledge graphs) and unstructured data (text embeddings), enabling a holistic comprehension of the dataset and facilitating the retrieval of relevant information across multiple contexts. By combining these approaches, SKETCH can perform multi-hop reasoning and retrieve contextually relevant information even if it is dispersed throughout the corpus. We evaluate SKETCH on diverse datasets, including QuALITY [3; 34], QAER [2; 8], NarrativeQA [1; 24], and an Italian Cuisine corpus, covering a range of domains and presenting various challenges. Our results demonstrate significant improvements in answer relevancy and context precision metrics compared to baseline methods. SKETCH outperforms traditional RAG approaches by maintaining context integrity and delivering more precise, contextually enriched responses, setting new benchmarks for RAG systems in handling large-scale discourse structures and complex queries across various domains.

---

[†]This work does not relate to the position at Amazon AI.
[‡]This work does not relate to the position at Meta.

## 2 Related Work

### 2.1 Retrieval Augmented Generation

RAG is a technique that enhances the capabilities of large language models (LLMs) by integrating external knowledge sources into the generation process [9; 25]. This approach addresses several limitations inherent in LLMs, such as their reliance on static training data and the potential for generating outdated or inaccurate information. By incorporating real-time, domain-specific knowledge, RAG systems can provide more accurate, relevant, and contextually enriched responses. This method not only improves the factual accuracy of the generated content but also enhances interpretability and provenance tracking, which are often lacking in traditional LLMs [13].

### 2.2 Retrieval Strategies

Retrieval methods have evolved from traditional term-based techniques like TF-IDF [21] to advanced strategies utilizing large language models as retrievers [10; 23; 26; 31; 41]. Innovations such as Fusion-in-Decoder (FiD) [20], combining DPR and BM25, and RETRO [5], employing cross-chunked attention and chunkwise retrieval, represent significant advancements. However, many models still rely on conventional techniques like chunking text corpora and using BERT-based retrievers, which have limitations in capturing the semantic depth of text [28], often leading to context loss in technical or scientific documents [7; 29; 46].

RAPTOR (Recursive Abstractive Processing for Tree-Organized Retrieval) [36] addresses these limitations by constructing a hierarchical tree structure that recursively embeds, clusters, and summarizes text chunks using SBERT [35] and Gaussian Mixture Models (GMMs) [11; 14; 40].

#### 2.2.1 Semantic Chunking

Semantic chunking is a relatively new technique used to divide text into semantically meaningful units, significantly improving the efficiency and accuracy of information retrieval in RAG systems [22]. Unlike traditional chunking methods based on simple rules or statistical models, semantic chunking leverages the inherent meaning of the text. The process begins by splitting the text into individual sentences, which are then grouped with neighboring sentences based on a window size k, representing the number of sentences before and after the current sentence to form a window. Vector embeddings are calculated for each window, and the cosine distance between sequential windows is evaluated. Window-merging strategies, such as calculating the 95th percentile of distance differences to set a threshold T as mentioned in [22], are employed. When the cosine distance difference between sequential windows is within this threshold T, the windows are merged into one chunk, and this process repeats until the threshold is breached, resulting in reformed documents within the corpus.

### 2.3 Knowledge Graphs

Knowledge graphs are powerful tools for representing and reasoning over structured knowledge by modeling entities and their relationships in a graph structure, enabling integration of information from diverse sources and facilitating knowledge discovery [15]. They provide rich context for understanding and retrieving information [4], with roots in semantic networks and conceptual graphs. Developments in the Semantic Web and standards like RDF and OWL have enhanced their utility [19; 33], leading to prominent examples like DBpedia, Wikidata, and the Google Knowledge Graph. Knowledge graphs have applications in domains such as question answering, recommender systems [38], and natural language processing tasks [18]. Research has focused on knowledge graph construction [39], embedding [37], completion [43], and reasoning [44], as well as integrating them with deep learning models for enhanced performance and explainable AI.

An approach of adding a semantic theme to chunking and creating hybrid retrievers with combination of structured data through knowledge graphs and unstructured data through semantic chunks are new and introduced through SKETCH.

## 3 SKETCH

### 3.1 Overview

#### 3.1.1 Semantic Chunking

In SKETCH, Semantic Chunking is crucial for enhancing the retrieval process by ensuring that text is segmented into semantically coherent units. Unlike traditional chunking methods that may disrupt the flow of ideas by splitting text arbitrarily, our Semantic Chunking approach preserves the thematic integrity of the content. This means each chunk represents a complete and meaningful segment of information, which is essential for accurate semantic embedding and retrieval. By maintaining the semantic continuity within chunks, SKETCH ensures that important contextual cues are not lost, which often happens with naive splitting techniques. This is particularly important when dealing with complex queries that require a deep understanding of the content. Semantic Chunking lays a strong foundation for the unstructured retrieval component of SKETCH, enabling more precise matching between queries and relevant text segments.

Moreover, this approach complements the structured retrieval provided by Knowledge Graphs. While Knowledge Graphs capture the relationships between entities, Semantic Chunking ensures that the unstructured text associated with these entities remains contextually rich and semantically intact. Together, they enable SKETCH to perform more accurate and contextually relevant retrievals, effectively handling complex queries that span multiple contexts within the corpus. By integrating Semantic Chunking into the SKETCH framework, we enhance the system's ability to understand and process large volumes of text, leading to significant improvements in retrieval accuracy and overall system performance.

#### 3.1.2 Knowledge Graphs

In SKETCH, Knowledge Graphs (KGs) are integral to enhancing the retrieval process by providing a structured representation of entities and their interrelationships within the corpus. We use LLM to derive the main subject or entities from a text snippet and then KGs to represent entities as nodes and their relationships as edges. In SKETCH, entity refers to the main subject that is under discussion in a sentence whereas the edges are the relationship that they have to other subjects in that sentence.

By encoding relationships between entities, KGs provide additional context that is not easily captured by text embeddings alone. This enriched context helps in understanding complex queries and retrieving more accurate information. When a user submits a query, SKETCH employs the KG for structured retrieval. We perform Named Entity Recognition (NER) [42] on the query using GPT-4 to extract relevant entities. These entities correspond to nodes in the KG. We then construct cypher queries to traverse the KG and retrieve pertinent nodes and their relationships based on the extracted entities. More specifically, KGs enable multi-hop reasoning, allowing the system to traverse multiple relationships to infer new information. When it comes to multi-context questions, the multi-hop feature of KG's fits perfectly in helping retrieve the required information that is potentially missed out during naive or even semantic chunking due to the distance between the texts in the corpus. This capability is particularly useful for answering complex, multi-faceted queries that require synthesizing information from various sources.

#### 3.1.3 Rationale for Combining Semantic Chunking and Knowledge Graphs

The innovative fusion of Semantic Chunking and Knowledge Graphs in SKETCH marks a significant leap forward in the field of Retrieval-Augmented Generation (RAG) systems. This strategic integration addresses core challenges in information retrieval, enabling SKETCH to deliver unprecedented accuracy and depth in handling complex queries across large corpora.

Semantic chunking ensures that text chunks are semantically coherent for a specific entity, while Knowledge Graphs (KGs) provide structured context about the relationships between entities. KGs enable multi-hop reasoning, which helps address complex queries that require integrating information from multiple distant sources. Addi-tionally, Knowledge Graphs offer clear traceability of information, making it easier to understand the flow of retrieval. The combination of Semantic Chunking and Knowledge Graphs creates a synergistic effect that amplifies the strengths of each approach while mitigating their individual limitations. This integration allows SKETCH to:

- **Maintain Semantic Integrity**: Ensuring that each chunk represents a semantically coherent unit reduces the risk of misinterpretation and enhances the quality of embeddings used for retrieval.

- **Enable Complex Reasoning**: Knowledge Graphs empower SKETCH to navigate the intricate web of relationships between entities, facilitating the retrieval of information that requires understanding multiple layers of context.

- **Enhance Retrieval Accuracy**: The dual approach ensures that both the depth (through semantic coherence) and breadth (through relational mapping) of information are captured, leading to more accurate and relevant retrieval results.

When merging results, SKETCH prioritizes semantic alignment: tokens that appear in both structured and unstructured contexts are treated as confirmation signals, reinforcing their relevance and importance. Therefore, integrating Semantic Chunking and Knowledge Graphs offers a more holistic context for each entity, enhancing the overall comprehension of the complete dataset.

## 3.2 Approach and Reproducibility

Refer Figure 2 to understand the architecture of SKETCH and how documents are pre-processed and embeddings are created and finally how user queries are processed through hybrid retrievers.

### 3.2.1 Indexing

**3.2.1.1 Document Loading and Initialization:** We load the dataset which serves as the initial corpus for indexing and retrieval. Additionally, two separate text splitters are initialized: a semantic text splitter and a recursive character text splitter. These splitters are responsible for dividing the text into meaningful chunks for further processing.

**3.2.1.2 Semantic Text Splitting:** The loaded documents are processed through a semantic text splitter. This segments the text based on semantic content rather than arbitrary lengths like paragraphs or sentences, ensuring each segment maintains thematic consistency. A new set of documents are created, where each document chunk represents a coherent semantic unit. This step is crucial for preserving the context and meaning within each chunk, enhancing the quality of information retrieval.

**3.2.1.3 Recursive Text Splitting:** The semantically segmented documents are further processed using a recursive character text splitter. This splitter divides the text into chunks of 100 tokens with a overlapping window of 16 tokens, ensuring that even large documents are broken down into smaller, more manageable pieces. By maintaining a chunk size of 100 tokens, the recursive text splitter ensures that the chunks remain contextually coherent, preventing the loss of important information that might occur if sentences were instead arbitrarily cut off.

**3.2.1.4 Embedding and Vector Store:** The embeddings generated after the recursive text splitting process are stored in a vector database, FAISS. These embeddings represent the semantic meaning of the text chunks and are used for efficient similarity-based retrieval during the querying phase.

**3.2.1.5 Knowledge Graph:** The initial set of documents is converted into graph documents. This involves parsing the text and identifying entities, attributes, and relationships, which are then structured into a graph format. Using the graph documents, a comprehensive Knowledge Graph (KG) is constructed. The KG captures the intricate relationships and connections between various entities, providing a structured representation of the information contained within the documents. Refer Figure 3 to understand the KG representation for the Italian Cuisine dataset. Each node here represents the entities that are retrieved from the Italian Cuisine corpus and their relationships are represented here as edges. Each node stores the smaller context of the corpus.

### 3.2.2 Querying

**3.2.2.1 Structured Retriever:** The first step in the structured retriever is Named Entity Recognition (NER). We identify all the plausibe entities present in the user query and create a cypher query based out of it, treating each entity as a node. We perform NER by passing the user query through GPT-4 and extracting a list of all entities in the sentence. The cypher query is designed to access and retrieve the relevant nodes and their relationships from the KG based on the extracted entities and relationships from the user's query.

**3.2.2.2 Unstructured Retriever:** This component queries the vector embeddings of the text chunks created during the recursive text splitting phase. By leveraging similarity-based retrieval technique, cosine similarity, the unstructured retriever can identify and retrieve the most relevant text chunks based on their semantic similarity to the user's query.

**3.2.2.3 Hybrid Retrieval:** The system combines the results from both the structured and unstructured retrievers to create a comprehensive and contextually rich retrieval mechanism. The retrieved results from these two components are then combined, forming a unified context that is subsequently fed to the Large Language Model (LLM) for generating answers to the user's queries. By leveraging the contextual coherence of semantically meaningful text chunks and the structured relationships captured in the Knowledge Graph, the system can provide more accurate and relevant information to the LLM, ultimately improving the quality of the generated responses.

## 3.3 Datasets

We evaluate the performance of SKETCH using four diverse datasets that present various challenges: a small "Italian Cuisine and Heritage" dataset and three large-scale datasets—QuALITY, QASPER, and NarrativeQA. These datasets test SKETCH's ability to handle long documents, multi-context questions, domain-specific knowledge, and multi-hop reasoning.

The Italian Cuisine dataset, consisting of 6,000 tokens across three text files, serves as an initial testbed for our methodology. We generated multi-context questions and ground truth data using the RAGAS framework and evaluated the approaches using RAGAS metrics: answer_relevancy, faithfulness, context_precision, and context_recall.

QuALITY [3; 34] is a multiple-choice question answering dataset designed for long document comprehension, with passages averaging 5,000 tokens. It tests the system's ability to process lengthy texts and answer deep comprehension questions that require understanding the entire passage. We used the training set, containing approximately 2,090 entries, for our experiments.

QASPER [2; 8] focuses on question answering over scientific papers, comprising 5,049 questions on 1,585 NLP papers. The questions cover methodology, results, and conclusions, requiring navigation through complex scientific texts and multi-hop reasoning to arrive at correct answers. We evaluated our approaches on the validation set, which contains 281 entries with questions, answers, and ground truths.

NarrativeQA [1; 24] involves question answering on long-form narrative texts, including 1,567 stories (books and movie scripts) and 46,765 question-answer pairs. The dataset challenges systems to understand and reason about complex narratives involving character relationships, plot developments, and thematic elements. Answers are free-form, allowing for nuanced and detailed responses. We used the validation set, containing approximately 3,460 entries, for our comparisons.

Table 1: Performance Comparison of Different Approaches against Italian Cuisine Dataset

| Approach | Answer Relevancy | Faithfulness | Context Precision | Context Recall | F1 Score |
|---|---|---|---|---|---|
| Naive RAG | 0.61 | **1.00** | 0.81 | **0.88** | 0.84 |
| Semantic | 0.84 | 0.86 | 0.92 | 0.83 | **0.87** |
| KG | 0.94 | 0.21 | 0.77 | 0.33 | 0.46 |
| RAPTOR | 0.75 | 0.73 | 0.38 | 0.71 | 0.50 |
| **SKETCH** | **0.94** | 0.87 | **0.99** | 0.72 | 0.83 |

## 4 Results and Discussion

We evaluated the performance of SKETCH across four diverse datasets: the "Italian Cuisine" dataset, QuALITY, QASPER, and NarrativeQA. Our analysis focused on comparing the effectiveness of SKETCH in enhancing retrieval accuracy and contextual relevance against existing approaches, including Naive RAG, RAPTOR, Semantic-only, and KG-only methods. To ensure a comprehensive evaluation, we assessed four key RAGAS metrics:

answer_relevancy, faithfulness, context_precision, and context_recall. GPT-3.5-turbo-16k served as the evaluation judge for all datasets, ensuring consistency in assessing the quality of responses generated by each method.

The results, visualized in Figure 2, clearly demonstrate SKETCH's significant advantage over other models in retrieval performance. SKETCH consistently performed better across multiple datasets, showing particular strength in answer relevancy, context precision metrics, making it a superior choice for accurate and contextually rich retrieval. Each plot in Figure 2 contrasts SKETCH with baseline models, providing a detailed visual representation of how SKETCH excels in various metrics, further emphasizing its robustness and balanced capability compared to the other tested approaches.

## 4.1 Italian Cuisine Dataset

The "Italian Cuisine" dataset, consisting of 6,000 tokens distributed across three text files, served as an ideal testbed to assess the core capabilities of SKETCH in a controlled environment. This dataset enabled us to evaluate SKETCH's performance in retrieving accurate and contextually relevant information. We generated a set of 9 multi-context questions that have answers spanning across different parargraphs in three files using the RAGAS framework to facilitate a comprehensive performance comparison.

- **Controlled Complexity for Initial Testing**: The dataset encompasses a variety of topics within the domain of Italian cuisine, such as regional dishes, traditional ingredients, culinary techniques, and cultural heritage. This diversity within a confined scope allows us to test SKETCH's ability to handle multi-faceted queries that require integrating information from different parts of the text.

- **Multi-Context Retrieval Challenges**: By generating a set of 9 multi-context questions using the RAGAS framework, we designed queries whose answers span across different paragraphs and even across multiple files.

- **Domain Diversity**: Including a dataset from a different domain ensures that our evaluation of SKETCH covers a broader spectrum of content types. While the other datasets—QuALITY, QASPER, and NarrativeQA—are centered around long-form narratives and scientific papers, the Italian Cuisine dataset represents a domain with unique characteristics.

- **Complexity Without Overhead**: The dataset strikes a balance between complexity and computational efficiency. It is rich enough to present significant retrieval challenges but small enough to allow for rapid iteration and testing. This is particularly beneficial during the development phase, where quick feedback loops are essential.

Table 1 presents the comparative metrics for the Italian Cuisine dataset across all five retrieval approaches: Naive RAG, RAPTOR, SKETCH, Semantic-only, and KG-only. SKETCH demonstrated clear superiority, achieving the highest Answer Relevancy score of 0.94, comparable only to the KG-only approach. However, SKETCH outperformed KG in Faithfulness (0.87 vs. 0.21) and Context Recall (0.72 vs. 0.33), highlighting SKETCH's ability to maintain context consistency more effectively. Compared to Naive RAG, SKETCH delivered a 54.1% improvement in Answer Relevancy, and it also exceeded RAPTOR by 26%. On the Context Precision metric, SKETCH outperformed RAPTOR by 160% and achieved an 22% higher score than Naive RAG. Although SKETCH's context-F1 score (0.83) was competitive, the Semantic-only approach slightly outperformed it in this metric (0.87), suggesting that while SKETCH excels in most areas, there may be room for further optimization in balancing precision and recall.

Figures 4, 5, 6, 7 and 8 illustrate the performance heatmaps for Naive RAG, RAPTOR, Semantic-only, and KG-only and SKETCH approaches, respectively. These heatmaps provide a deeper understanding of how each approach handled the 9 questions in the test set. SKETCH consistently demonstrated a higher level of answer relevancy and contextual accuracy compared to the other methods, emphasizing its capability to deliver precise and contextually rich responses. The additional heatmaps for Semantic-only and KG-only approaches further illustrate that SKETCH excels in providing well-rounded

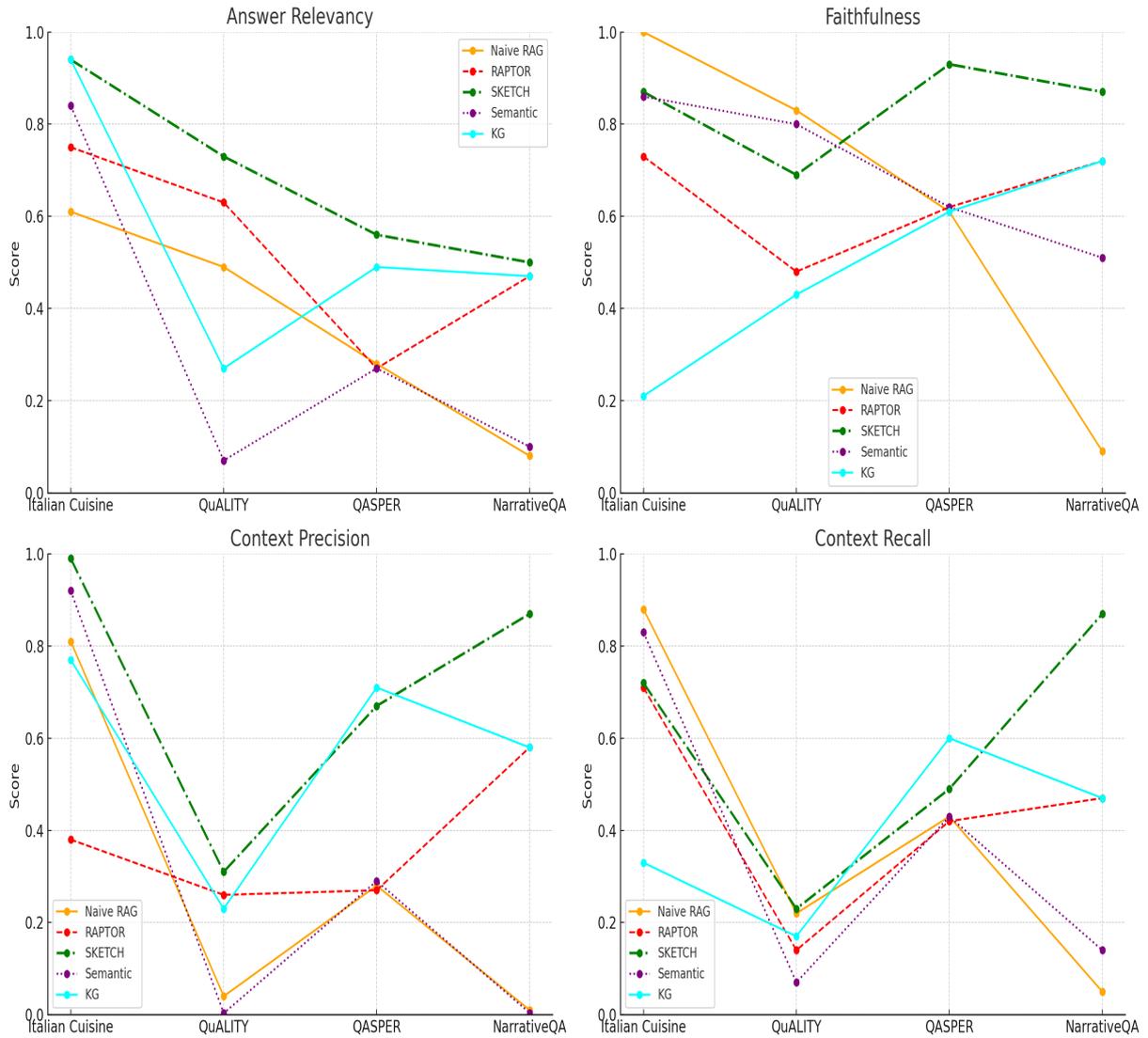

Figure 1: **RAGAS metrics against all datasets and approaches**

retrieval results, surpassing the limitations seen in isolated semantic or structured retrieval methods.

## 4.2 QuALITY Dataset

The QuALITY validation dataset, designed for long-form document comprehension, provided a challenging evaluation ground for retrieval-augmented systems. Despite the complexity of extended passages, SKETCH outperformed all baseline methods, demonstrating its capability to retrieve and integrate context effectively. Table 2 compares SKETCH and other approaches across key performance metrics.

For Answer Relevancy, SKETCH achieved a score of 0.73, representing a 49% improvement over Naive RAG (0.49) and a 15.87% improvement over RAPTOR (0.63). The Semantic-only and KG-only approaches scored significantly lower at 0.07 and 0.27, respectively. In terms of Faithfulness, SKETCH recorded 0.69, while Naive RAG performed better with 0.83. Although Naive RAG had higher faithfulness, SKETCH's strong performance across other metrics gave it an overall edge. The Semantic-only and KG-only approaches scored 0.80 and 0.43, respectively. This demonstrates SKETCH's effectiveness in synthesizing relevant information from different parts of long passages.

Regarding Context Precision, SKETCH achieved 0.31, a 675% improvement over Naive RAG's 0.04 and a 19.23% improvement over RAPTOR's 0.26. SKETCH also surpassed the KG-only approach, which scored 0.23. For Context Recall, SKETCH recorded 0.23, comparable to Naive RAG's 0.22. RAPTOR, Semantic-only, and KG-only scored 0.14, 0.07, and 0.17, respectively. Compared to other approaches, SKETCH's superior context based F1 score underscores its ability to provide both accurate and contextually consistent retrieval, even in the face of complex, long-form content. These results illustrate SKETCH's superior performance across key metrics, reinforcing its effectiveness as an integrated retrieval mechanism for comprehending long and complex documents over other baseline approaches and showcasing SKETCH's advantage in accurately pinpointing relevant context.

Table 2: Performance Comparison of Different Approaches against QuALITY Dataset

| Approach | Answer Relevancy | Faithfulness | Context Precision | Context Recall | F1 score |
|---|---|---|---|---|---|
| Naive RAG | 0.49 | **0.83** | 0.04 | 0.22 | 0.07 |
| RAPTOR | 0.63 | 0.48 | 0.26 | 0.14 | 0.18 |
| Semantic | 0.07 | 0.80 | 0.003 | 0.07 | 0.01 |
| KG | 0.27 | 0.43 | 0.23 | 0.17 | 0.20 |
| **SKETCH** | **0.73** | 0.69 | **0.31** | **0.23** | **0.26** |

### 4.3 QASPER

The QASPER dataset, consisting of scientific papers, presented a challenging environment due to its complexity and technical nature. Despite these challenges, SKETCH demonstrated strong capabilities, outperforming all baseline methods across several key metrics, as shown in Table 3.

For Answer Relevancy, SKETCH achieved a score of 0.56, significantly surpassing Naive RAG's 0.28 a 100% improvement and outperforming RAPTOR's 0.27 by 107.41%. The Semantic-only and KG-only approaches scored 0.27 and 0.49, respectively. On the Faithfulness metric, SKETCH scored an impressive 0.93, exceeding Naive RAG (0.61) by approximately 52.46% and RAPTOR (0.62) by 50%. Regarding Context Precision, SKETCH achieved 0.67, outperforming Naive RAG's 0.28 by 139.29% and RAPTOR's 0.27 by 148.15%. While the KG-only approach achieved a slightly higher precision at 0.71, SKETCH's performance was more balanced across all metrics.

For Context Recall, SKETCH recorded a score of 0.49, improving over Naive RAG's 0.43 by 13.95%. Although the KG-only approach scored higher at 0.60, SKETCH demonstrated superior balance, excelling in relevancy, faithfulness, and precision. Although KG-only achieves the highest F1 score (0.65), SKETCH still maintains a strong F1 performance at 0.57, reflecting its balanced effectiveness. These results indicate that SKETCH consistently outperforms other approaches in the QASPER dataset, emphasizing its strength in providing accurate, faithful, and contextually precise retrieval, particularly in challenging scientific content.

Table 3: Performance Comparison of Different Approaches against QASPER Dataset

| Approach | Answer Relevancy | Faithfulness | Context Precision | Context Recall | F1 score |
|---|---|---|---|---|---|
| Naive RAG | 0.28 | 0.61 | 0.28 | 0.43 | 0.34 |
| RAPTOR | 0.27 | 0.62 | 0.27 | 0.44 | 0.33 |
| Semantic | 0.27 | 0.62 | 0.29 | 0.43 | 0.35 |
| KG | 0.49 | 0.61 | **0.71** | **0.60** | **0.65** |
| **SKETCH** | **0.56** | **0.93** | 0.67 | 0.49 | 0.57 |

### 4.4 NarrativeQA

The NarrativeQA dataset posed unique challenges requiring deep narrative understanding, including handling plot development and character interactions across extended text passages. SKETCH demonstrated a notable advantage in addressing these complexities compared to baseline methods, as detailed in Table 4.

For Answer Relevancy, SKETCH achieved a score of 0.50, significantly surpassing Naive RAG's 0.08 (a 525% improvement) and slightly outperforming RAPTOR and KG-only, both at 0.47. The Semantic-only approach scored 0.10. In Faithfulness, SKETCH recorded 0.87, outshining Naive RAG (0.09) by 866.67% and improving over RAPTOR and KG-only, both at 0.72, by 20.83%. Regarding Context Precision, SKETCH achieved 0.51, better than Naive RAG's 0.10 and RAPTOR's

0.30. Although KG-only slightly outperformed SKETCH with 0.58, SKETCH's overall balanced performance ensured its superiority. For Context Recall, SKETCH scored 0.46, significantly higher than Naive RAG's 0.05 (an 800% improvement) and surpassing RAPTOR (0.16) and Semantic-only (0.14), with KG-only slightly ahead at 0.47.

These results indicate that SKETCH consistently outperformed baseline methods in answer relevancy and faithfulness. While KG-only showed slightly better scores in context precision and recall, SKETCH's balanced excellence across all metrics made it the most effective approach for retrieving and synthesizing narrative content. This demonstrates SKETCH's ability to provide contextually rich and accurate retrieval in the NarrativeQA dataset, confirming its superiority in comprehending and generating answers from complex narrative passages.

Table 4: Performance Comparison of Different Approaches against NarrativeQA Dataset

| Approach | Answer Relevancy | Faithfulness | Context Precision | Context Recall | F1 score |
|---|---|---|---|---|---|
| Naive RAG | 0.08 | 0.09 | 0.10 | 0.05 | 0.07 |
| RAPTOR | 0.10 | 0.46 | 0.30 | 0.16 | 0.21 |
| Semantic | 0.10 | 0.51 | 0.004 | 0.14 | 0.01 |
| KG | 0.47 | 0.72 | **0.58** | **0.47** | **0.52** |
| **SKETCH** | **0.50** | **0.87** | 0.51 | 0.46 | 0.48 |

## 5 Conclusion and Limitations

This research introduced SKETCH, an innovative methodology that enhances Retrieval-Augmented Generation (RAG) systems by combining semantic chunking with knowledge graphs. By integrating structured and unstructured data, SKETCH addresses the limitations of traditional retrieval methods, such as context loss in large datasets and limited comprehension of complex queries. Our experiments on diverse datasets—including Italian Cuisine, QuALITY, QASPER, and NarrativeQA—demonstrate SKETCH's superior ability to maintain context integrity and generate highly relevant responses, consistently outperforming baseline methods like Naive RAG, RAPTOR, Semantic-only, and KG-only approaches.

SKETCH achieved remarkable improvements across key metrics, showcasing its adaptability to both short and long documents and its effectiveness in navigating complex texts. For instance, on the Italian Cuisine dataset, it achieved an answer relevancy score of 0.94 and context precision of 0.99, improving over Naive RAG by 54.1% and 22.2%, respectively. On the QuALITY dataset, SKETCH improved answer relevancy by 49%, and on the QASPER dataset, it doubled the answer relevancy score over Naive RAG while increasing context precision by 139.29%. Even on the challenging NarrativeQA dataset, SKETCH delivered balanced performance with a 525% improvement in answer relevancy and an 866.67% increase in faithfulness over Naive RAG. While KG achieved the highest F1 score (0.52), SKETCH closely followed with 0.48, demonstrating its balanced effectiveness despite the complexity of narrative content. These findings confirm that SKETCH's combined approach offers a robust framework for advancing RAG systems, setting new benchmarks in accuracy, context comprehension, and cross-domain xapplicability, and paving the way for future advancements in natural language processing.

*Limitations* While SKETCH significantly improved Answer Relevancy and Context Precision, certain limitations remain. Faithfulness, though better than most baselines, still lags behind Naive RAG on QuALITY (0.69 vs. 0.83, a 16.9% shortfall). Scalability and cost are also concerns, as constructing large-scale knowledge graphs is labor-intensive, and relying on paid LLMs like GPT-4 for semantic evaluation increases expenses. Furthermore, SKETCH's dependence on GPT models for query parsing and RAGAS evaluation can introduce errors and variance due to sampling randomness, prompt sensitivity, and occasional hallucinations, potentially affecting reproducibility. Employing sampling strategies (e.g., multiple runs) and aggregating judgments could help mitigate these issues. Future work should focus on reducing computational costs, refining knowledge graph construction, and improving metrics like Context Recall and Faithfulness to achieve more consistent, high-quality, and stable retrieval results.

## Acknowledgments

This research was supported by an OpenAI Research Grant through Researcher Access Program.

## A  Appendix

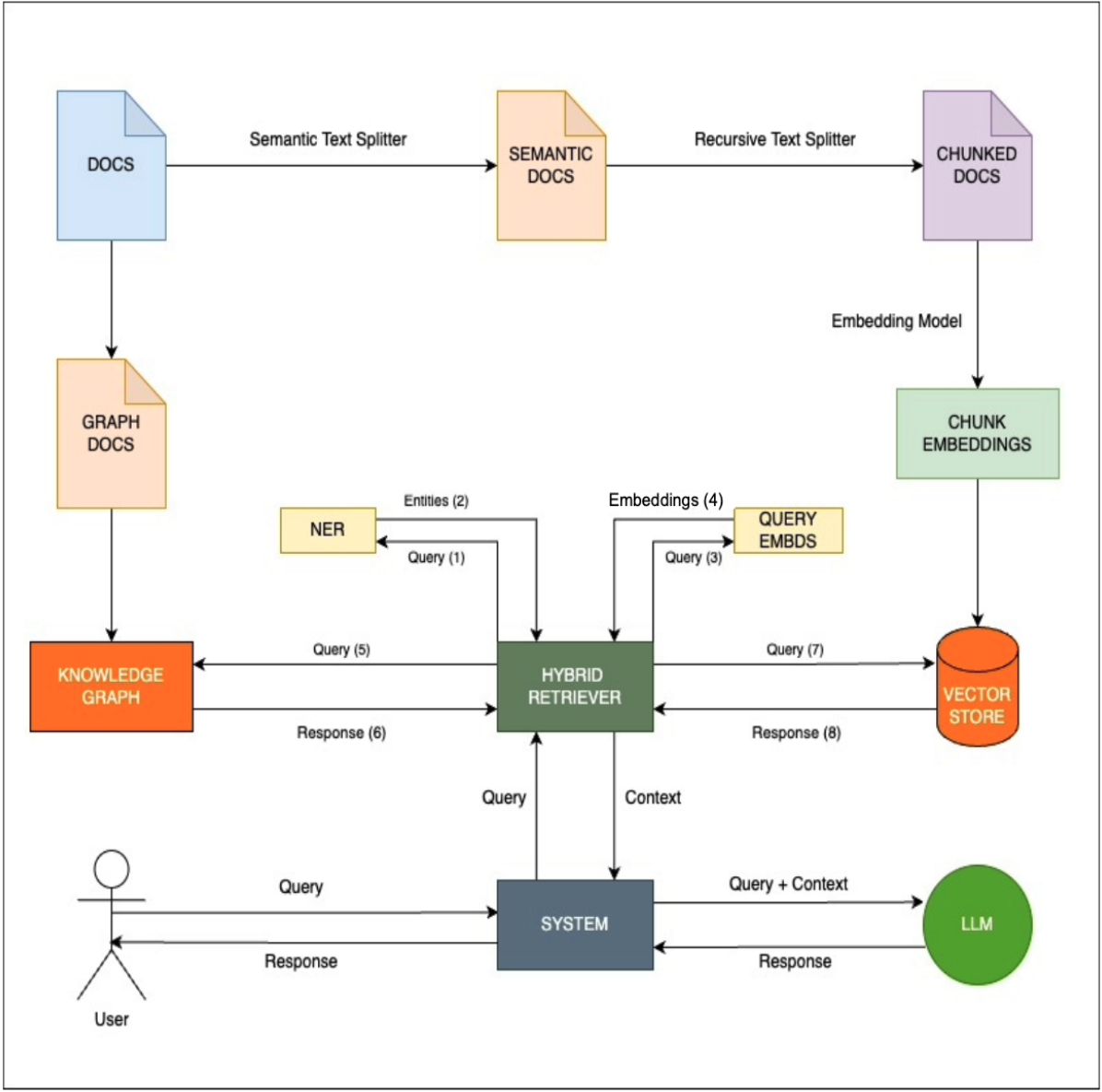

Figure 2: Architecture of SKETCH with a Hybrid Retriever combining structured and unstructured retrievers

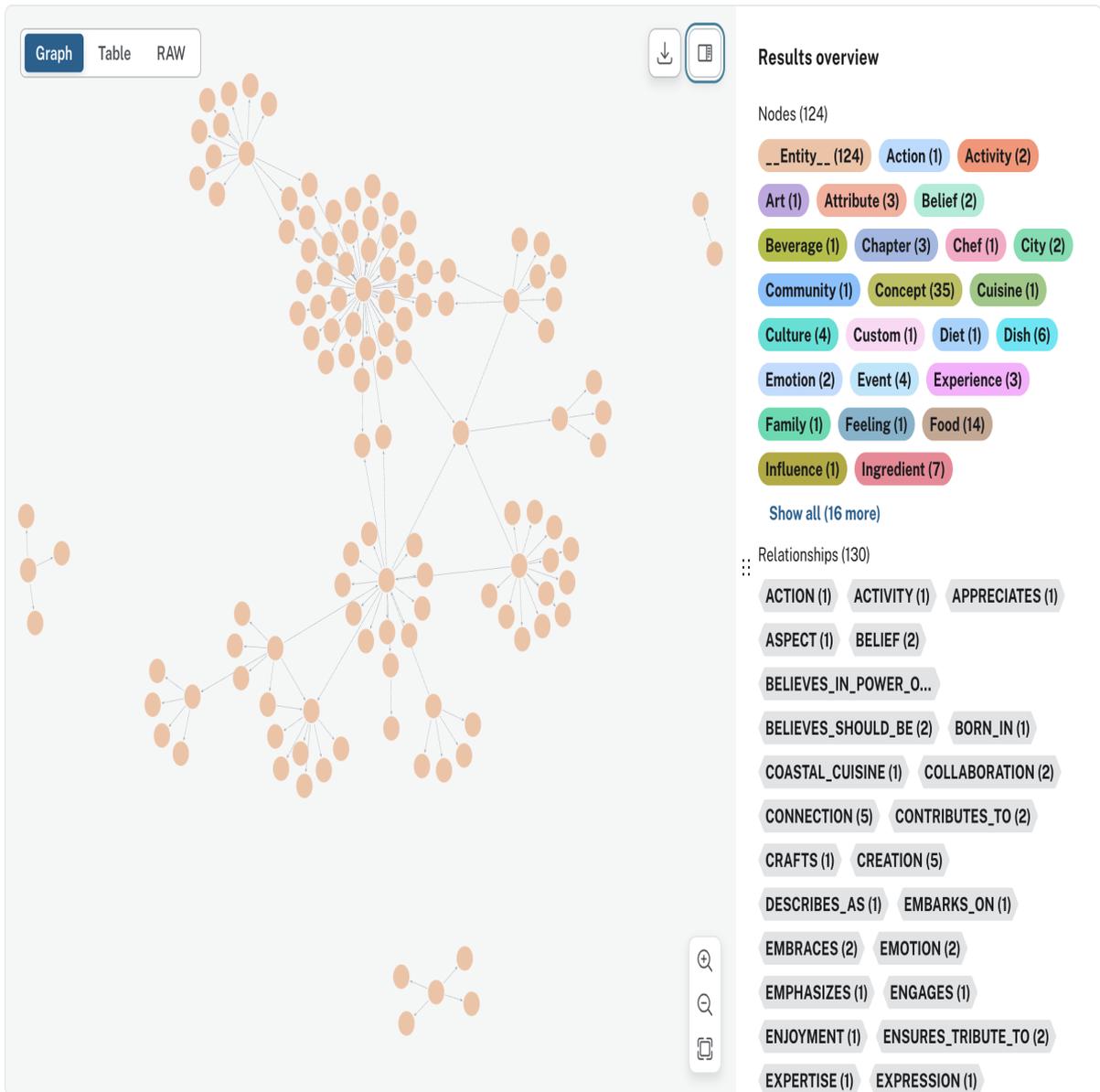

Figure 3: Italian Cuisine KG Representation

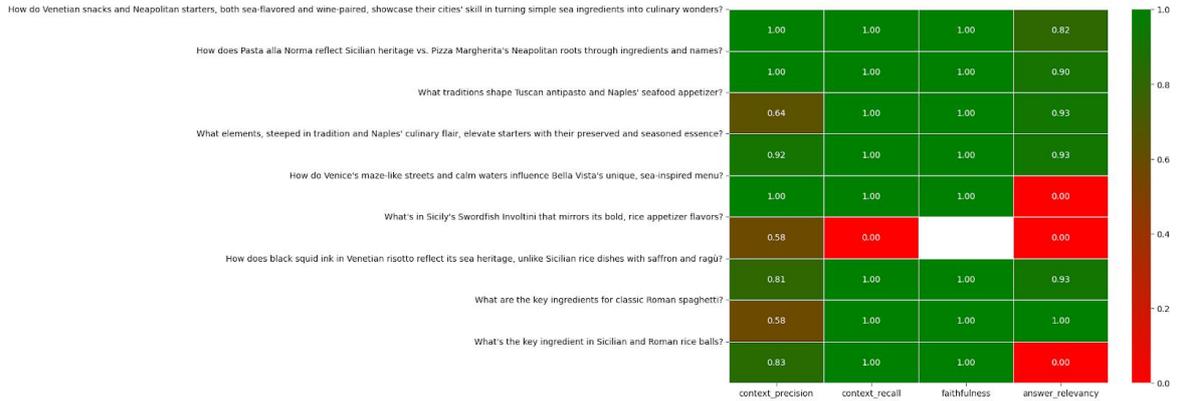

Figure 4: Naive RAG Italian Cuisine Performance Heatmap

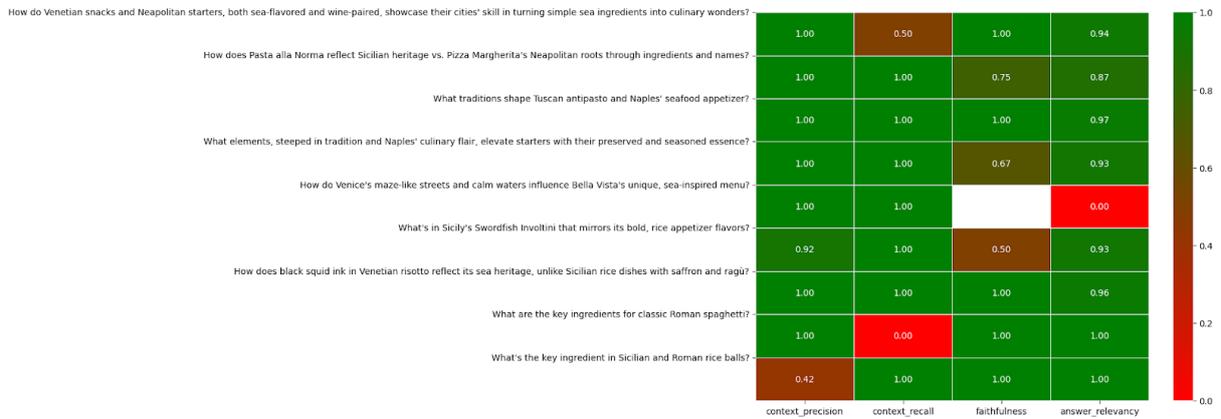

Figure 5: Semantic only Italian Cuisine Performance Heatmap

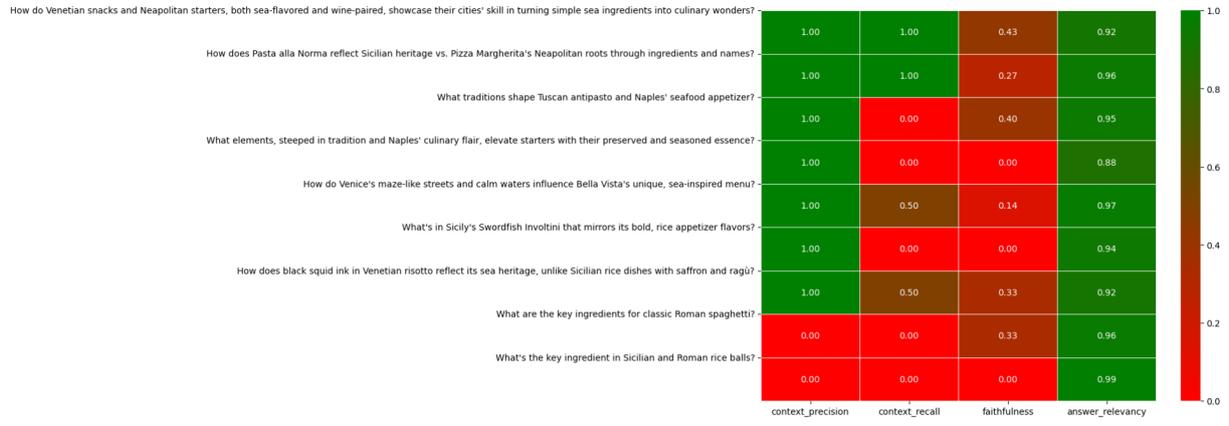

Figure 6: KG only Italian Cuisine Performance Heatmap

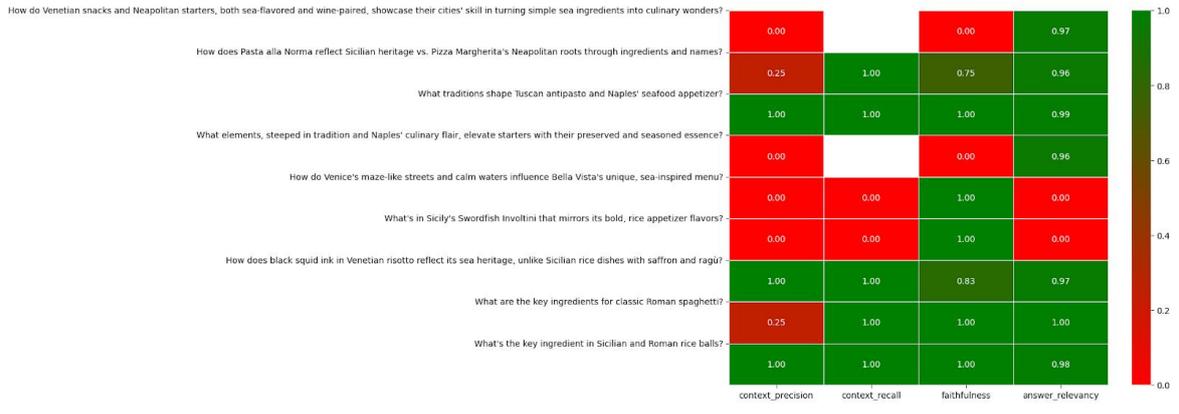

Figure 7: RAPTOR Italian Cuisine Performance Heatmap

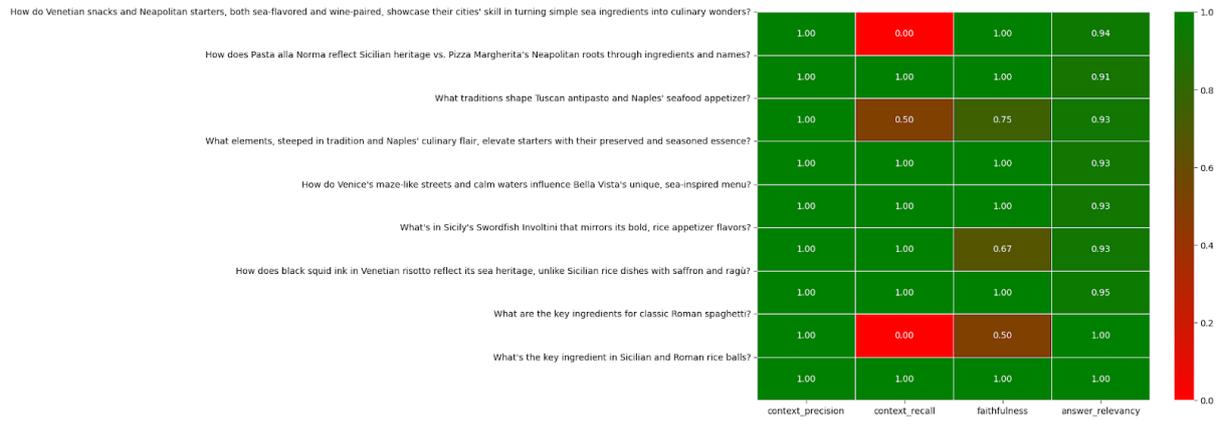

Figure 8: SKETCH Italian Cuisine Performance Heatmap